\pdfoutput=1

\documentclass[11pt]{article}

\usepackage{acl}

\usepackage{times}
\usepackage{latexsym}

\usepackage[T1]{fontenc}

\usepackage[utf8]{inputenc}

\usepackage{graphicx} 
\usepackage{microtype}
\usepackage{multirow}
\usepackage{color}
\usepackage{makecell}
\usepackage{amssymb}
\usepackage{amsmath}
\usepackage{algorithm}
\usepackage{algorithmic}
\title{S2F-NER: Exploring Sequence-to-Forest Generation for Complex Entity Recognition}

\author{Yongxiu Xu\textsuperscript{\rm 1,2}\thanks{Corresponding  author} ,
	Heyan Huang\textsuperscript{\rm 3},
	Yue Hu\textsuperscript{\rm 1,2} \\
  \textsuperscript{\rm 1} Institute of Information Engineering Chinese Academy of Sciences, Beijing, China \\ 
   \textsuperscript{\rm 2} School of Cyber Security, University of Chinese Academy of Sciences, Beijing, China \\
  \textsuperscript{\rm 3} School of Computer Science and Technology, Beijing Institute of Technology, Beijing, China  \\
  \texttt{ xuyongxiu@iie.ac.cn, hhy63@bit.edu.cn, huyue@iie.ac.cn} \\}

\begin{document}
\maketitle
\begin{abstract}
Named Entity Recognition (NER) remains challenging due to the complex entities, like nested, overlapping, and discontinuous entities. Existing approaches, such as sequence-to-sequence (Seq2Seq) generation and span-based classification, have shown impressive performance on various NER subtasks, but they are difficult to scale to datasets with longer input text because of either exposure bias issue or inefficient computation. In this paper, we propose a novel \textbf{S}equence-to-\textbf{F}orest generation paradigm, \textbf{S2F-NER}, which can directly extract entities in sentence via a Forest decoder that decode multiple entities in parallel rather than sequentially. Specifically, our model generate each path of each tree in forest autoregressively, where the maximum depth of each tree is three (which is the shortest feasible length for complex NER and is far smaller than the decoding length of Seq2Seq). Based on this novel paradigm, our model can elegantly mitigates the exposure bias problem and keep the simplicity of Seq2Seq. Experimental results show that our model significantly outperforms the baselines on three discontinuous NER datasets and on two nested NER datasets, especially for discontinuous entity recognition\footnote[1]{We will release our code and data for future research.}.
\end{abstract}

\section{Introduction}
 \label{intro: Introduction}
Named entity recognition is capable of extracting the textual information consisting of named entity mentions and types from unstructured text, has appealed to many attention increasingly. This is largely due to its crucial role in many downstream natural language processing (NLP) tasks, such as relation extraction~\cite{eberts2020span}, event extraction~\cite{wadden2019entity} and question answering~\cite{cao2019bag}.

\begin{figure}[t]
	\centering
	\includegraphics[width=0.9\columnwidth]{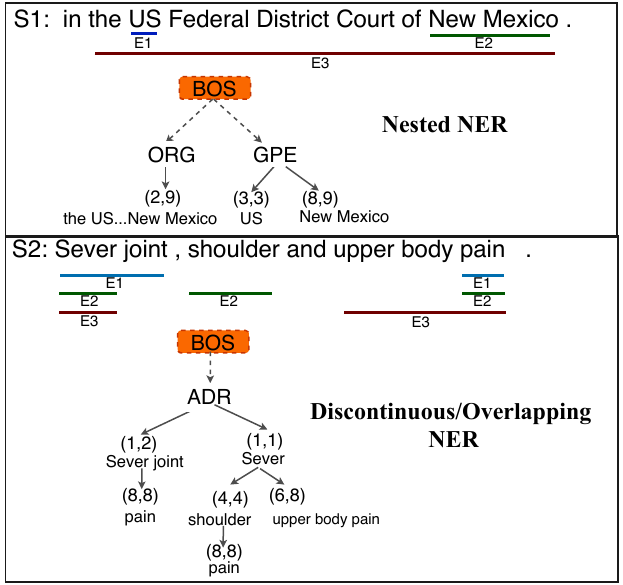}
	\caption{Examples involving three kinds of complex NER subtasks and the corresponding mainstream solution of our proposed method. Entities are highlighted with colored underlines. Note that ``BOS" is a dummy root of a whole sentence.} \label{fig1}
\end{figure}

Traditionally, the NER system formulates NER as a sequence labeling problem by conditional random field (CRF) based models~\cite{lafferty2001conditional}, which assigns a single label to each token based on two underlying assumptions: 1) entities do not overlap with each other, and 2) each entity consist of continuous sequence of words. While this sequence labeling formulation has dramatically advanced the NER task, it can only solve flat NER where a token can only belong to one entity, is difficult to handle complex NER~\cite{dai2018recognizing} where entity mentions may be nested, overlapped or discontinuous in many practical scenarios. Consider the examples in Figure~\ref{fig1}. Both \textit{New Mexico} and \textit{the US Frederal District Court of New Mexico} are two continuous entities, but they are nested with each other. \textit{Sever joint pain} is a discontinuous entity involving two fragments. Both \textit{Sever joint pain} and \textit{Sever shoulder pain} are two discontinuous entities, but they are overlapped each other. Apparently, it is more challenging to extract complex entities than traditional standard NER (i.e., flat NER).

To the best our knowledge, only several approaches for recognizing both the nested and discontinuous entities simultaneously, has been proposed. Current state-of-the-art (SOTA) unified models can be categorized into two classes: span-based~\cite{li2021span} and Seq2Seq-based~\cite{hang2021a} models. Span-based model~\cite{li2021span} exploits span-level classification based on span representation and pairwise scoring techniques to determine whether two specific fragments involving a predefined relation (\texttt{Overlapping }or \texttt{Succession}). However, this approach suffers from high time/space complexity because it needs to enumerate all possible candidate text spans~\cite{luan2019general}. The Seq2Seq-based model ~\cite{hang2021a} is flexible to be adapted for complex NER using the pre-trained language model BART~\cite{lewis2020bart}. However, this approach may severely suffers from exposure bias~\cite{zhang2019bridging} due to the long autoregressive decoding length. Moreover, the Seq2Seq model enforces an unnecessary order on such unordered recognition task, thus the predefined entity sequential order makes the model prone to overfit to the frequent orders in the training set and poorly generalize to the unseen orders.  While above models have achieved reasonable performance on various NER subtasks, they are impractical for long sequences because of either inefficient computation or exposure bias. Overall, it is still challenging to recognize nested and discontinuous (potentially overlapping) entities concurrently using a single unified model\footnote[2]{Note that there certain words may be overlapped in discontinuous entity mentions, and thus extracting discontinuous entities also need to handle the overlapping mentions. Additionally, nested NER can be regard as a special case of overlapping NER.}.

To mitigate the aforementioned problems while keeping the simplicity of the Seq2Seq model, we recast the entity sequence in text into a forest consists of multiple entity tree where each node is a mention span except the root node which represents the entity type of each tree, as shown in Figure~\ref{fig1}. Note that, we introduce a dummy root of the forest, named``BOS", to fit in the generation paradigm. In this paper, we propose a novel unified model, named S2F-NER, to recognize nested and discontinuous (potentially overlapping) entities simultaneously. The three core insights behind S2F-NER are: 1) discontinuous NER can be seen as continuous NER if we only need to identify the first fragment of a discontinuous entity; 2) if one model can not fully perceive the semantic of the previous fragment of discontinuous entity, it will be unreliable to recognize the subsequent fragments. 3) for NER task, the output entities are essentially an unordered set while the words consisting of entities are ordered, which demonstrates that it is more suitable for modeling NER as a entity forest than a sequence of entities. 

Specifically, we establish our model using an encoder-decoder framework, where the decoder is equipped with the scratchpad attention mechanism~\cite{zhang2020minimize} and the biaffine-attention based span detector to effectively generate the mention span and ensure the span nodes in the same layer are unordered. Different from the Seq2Seq, the decoding length is limited to 3 because most entities (about 99.8\% in the datasets used in this paper) have no more than three fragments\footnote[3]{The datasets for discontinuous NER, i.e., CADEC, ShARe 13 and ShARe 14,we set the depth of the tree to be 3, while for nested NER, i.e., GENIA and ACE05, we set to 1.}. Owing to our new paradigm, our model can minimize the exposure bias of Seq2Seq model and largely reduce the time/space complexity. 
Our main contributions in this paper can be summarized as follows:
\begin{itemize}
	\item We propose a novel unified method, a sequence-to-forest generation, for complex NER tasks including nested NER and discontinuous (potentially overlapping) NER, which can directly extract entities in an end-to-end fashion.
	\item We present a novel Forest decoder which can remit the exposure bias issues while keeping the simplicity of the Seq2Seq model. Also, it can capture inter-dependencies among spans and types due to its nature as a sequential decision process.
	\item We conduct extensive experiments on multiple benchmark datasets including three discontinuous NER datasets and two nested NER datasets. The results show that our model outperforms or achieve comparable performance with baselines, which demonstrates the effectiveness of our model. 
\end{itemize}

\section{Related Work}
There have been several studies to investigate complex NER subtasks, which mainly fall into two categories: the \textbf{nested} NER~\cite{alex2007recognising,finkel2009nested,lu2015joint,muis2017labeling,ju2018neural,sohrab2018deep,fisher2019merge,strakova2019neural,ringland2019nne,zheng2019boundary,wang2020pyramid,shibuya2020nested,xu2021supervised}, breaks the first assumption of flat NER (as described in the \nameref{intro: Introduction}), and the \textbf{discontinuous} NER~\cite{tang2013recognizing,metke2016concept,dai2017medication,tang2018recognizing,wang2019combining,dai2020effective,hang2021a,li2021span} breaks the second assumption of flat NER, yet it comparatively less be studied. 
\subsection{Nested NER}
We call both of the entities involved as nested entities, where one entity mention is completely embed in the another. The nested structures between entities was first noticed by~\cite{kim2003genia}, who developed handcrafted rules to identify nested (or overlapping) mentions. With the emergence and development of deep representation learning, neural models has also been utilized in nested NER. One representative category is sequence labeling based models~\cite{wang2018neural,strakova2019neural} which aim to transform the nested structures into flat structures and then utilize the well-studied neural LSTM-CRF architecture~\cite{huang2015bidirectional}. Another appealing category is span-base models~\cite{zheng2019boundary,xia2019multi,wang2020pyramid,tan2020boundary} which classified all possible candidate spans based on span representations.  Though the above approaches have achieved great success in nested NER, however they are incapable of handling the discontinuous entities.

\subsection{Discontinuous NER}
Discontinuous NER~\cite{metke2016concept} is comparatively less studied than nested NER due to its more complex structures of discontinuous entities which contains at least one mention span (or fragment) and can potentially overlapped with other entities. Inspired by the success of tagging scheme in flat NER,~\cite{tang2013recognizing,metke2016concept,dai2017medication,tang2018recognizing} tried to extend the ``BIO" schemes by introducing new labels for continuous mention span shared by discontinuous entities or not, such as  ``BIOHD" and ``BIOHD1234". However, most of these methods suffers from the tag ambiguity problem due to the limited flexibility of the extended tag set. Currently,~\cite{dai2020effective} present a transition-based model by incrementally labeling the discontinuous span via a sequence transition actions. However, this model may suffer from the long-range dependency. Different from the above models, we propose a better solution to recognize all types of complex entities in a fully end-to-end manner.

\subsection{Joint Complex NER subtasks}
To the best of our knowledge, there are only several publications which has tested their model or framework on various complex NER datasets. These models mainly include  the span-level classification model~\cite{li2021span}, where complex entities are recognized by a span classification framework for all enumerated spans, and the Seq2Seq generative-based model~\cite{fei2021rethinking,li2021span}, where the decoder predicts a sequence of indexes of words (or spans) consisting of entities. However, both models are difficult to scale to datasets with longer input text because of either severe exposure bias issue of the Seq2Seq-based model or high time/space complexity of the span-based model. In this paper, we propose a novel paradigm, sequence-to-forest generation, to extract various complex entities directly, effectively and simultaneously .
\begin{figure*}[t]
	\centering
	\includegraphics[width=0.9\textwidth]{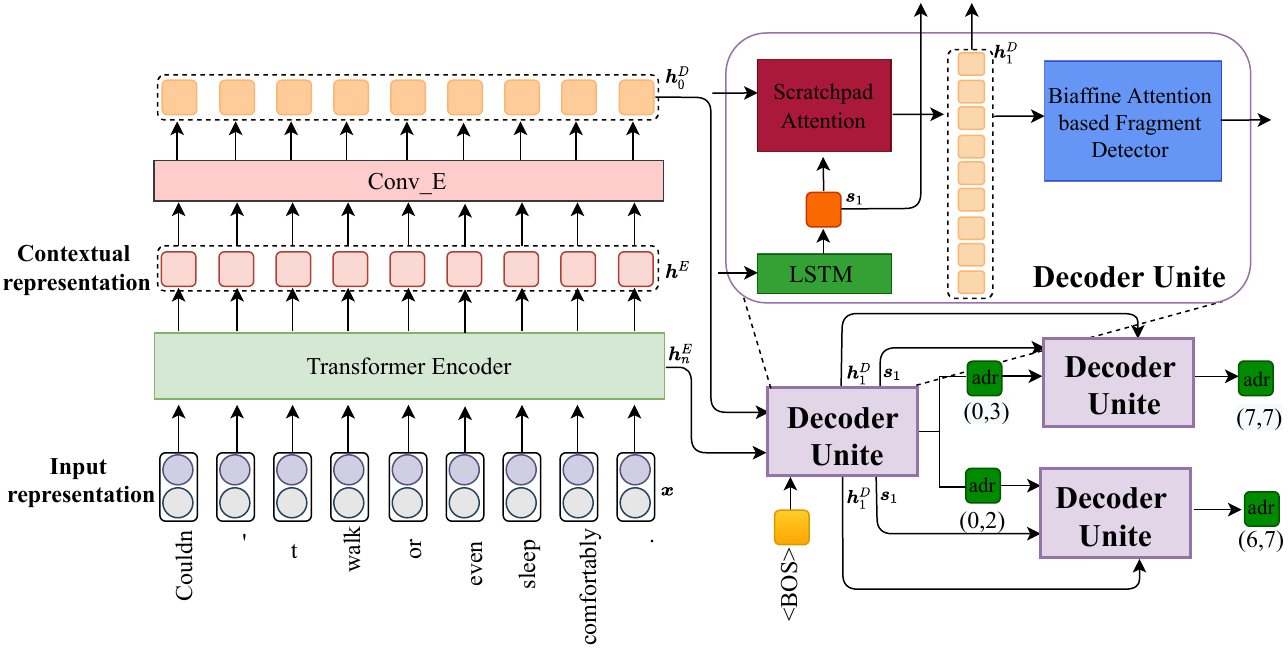}
	\caption{The overview of our model. ``<BOS>" is the predefined begin-of-sequence token and represents a dummy root of a whole sentence. In the decoder, ``adr" means the entity type.} 
	\label{fig2}
\end{figure*}
\section{Methodology}

\subsection{Method Overview}

We build our model using an encoder-decoder architecture, where the encoder comprised of a conventional Transformer encoder~\cite{devlin2019bert} and a novel Forest decoder which mainly contains two mechanism: scratchpad attention and biaffine attention mechanisms. The architecture of our model is illustrated in Figure ~\ref{fig2}. 

Our model aims to extract a set of entities that may have nested, overlapping and discontinuous structures in a natural language sentence. Given an input sentence $X=\{w_1,...,w_n\}$ with $n$ tokens, our model outputs a set of entities $Y=\{(f_{i_1:j_1},...,f_{i_m:j_m})^{k}\}$, where $k$ denotes the entity type and $(f_{i_1:j_1},...,f_{i_m:j_m})$ denotes a list of fragment spans that make up the entity, the subscript of spans e.g., $i_1:j_1$ (or $i_m:j_m$), indicates the starting and ending positions of the span in the input sequence. Hence,  the length of span list for continuous entity is 1, for discontinuous entity is greater than 1 and less than or equal to 3 (i.e., $1\textless m\leq 3$). Additionally, we design one special token, namely \texttt{BOS} (``begin-of-sequence"), which is always the beginning of the decoding and is considered as depth 0. We will introduce the model details in the following subsections.

\subsection{Feature Encoder }
The feature encoder aims to map input tokens to distributed semantic representations, consisting of a Transformer encoder~\cite{vaswani2017attention}  and a local feature convolutional layer.
For the input sentence $X=\{w_1,...,w_n\}$, we first obtain the input representation $\boldsymbol{x}_i$  for each token $w_i$. Each $\boldsymbol{x}_i$ is represented by concatenating two sources: a character-level token embedding $\boldsymbol{x}_i^c$ and a absolute position embedding $\boldsymbol{x}_i^p$, as follows:
\begin{equation}\label{eq1}
\boldsymbol{x}_i=[\boldsymbol{x}_i^{c};\boldsymbol{x}_i^{p}], i\in [1,n]
\end{equation}
where, a convolutional neural network (CNN) is utilized to obtain $\boldsymbol{x}_i^c$.

Then, we adopt Transformer encoder, which is known as a backbone structure of various pre-trained language models and has shown to be prominent on a wide range of NLP tasks, to encode contextual information into word representations $\boldsymbol{x}=(\boldsymbol{x}_i,...,\boldsymbol{x}_n)$. 
\begin{equation}\label{eq2}
\boldsymbol{h}_i^E,...,\boldsymbol{h}_n^E=\mathrm{Transformer\:Block}_L(\boldsymbol{x}_i,...,\boldsymbol{x}_n)
\end{equation}
where $L$ is the number of stacked Transformer blocks.

Then the output sequence $\boldsymbol{h}^E$ is fed into a one-dimensional convolutional neural layer to extract n-gram features at different positions of a sentence
and reduce the dimensions of the contextual representations.

\begin{equation}\label{eq3}
\boldsymbol{c}_i,...,\boldsymbol{c}_n=\mathrm{Conv\_E}(\boldsymbol{h}_i^E,...,\boldsymbol{h}_n^E)
\end{equation}

Here, the output sequence $\boldsymbol{c}_i,...,\boldsymbol{c}_n$ is further denoted as $\boldsymbol{h}_0^D$ which is used for the scratchpad attention of the decoder. 

\subsection{Forest Decoder}
The main difference between Forest decoder and the standard Seq2Seq decoder is that it generates unordered multiple entity spans and has far shorter decoding length (limited to 3). Each unite of our Forest decoder mainly consists of three components: backbone decoder (LSTM decoder), scratchpad attention mechanism and biaffine-attention based span detector. Next, we will detail each component.

\subsubsection{Backbone Decoder \label{decoder}}
We employ the LSTM~\cite{hochreiter1997long} neural network as the backbone decoder of our model. Given the input embedding $\boldsymbol{w}_t$ and the previous decoder hidden state $\boldsymbol{s}_{t-1}$, a LSTM cell is utilized to produce the current decoder hidden state $\boldsymbol{s}_t$, as follows:

\begin{equation}\label{eq4}
\boldsymbol{s}_t=\mathrm{LSTM}(\boldsymbol{w}_t,\boldsymbol{s}_{t-1})
\end{equation}
where the initial hidden state $\boldsymbol{s}_0$ is set to the final state of the encoder, i.e., $\boldsymbol{h}^E_n$. $t$ is the decoding step, for nested NER datasets, we set $t==1$ since all entities are continuous (i.e. only have one fragment), while for discontinuous NER datasets, we set $t \in \{1,2,3\}$.

For the input embedding $\boldsymbol{w}_t$, we initialize the $\boldsymbol{w}_1$ by $\boldsymbol{w}^{(BOS)}$ since the decoder always takes the special token \texttt{BOS} at the first-time step, use $\boldsymbol{w}^{type}_t$ and  $\boldsymbol{w}^{f_{i:j}}_t$ to represent entity type embedding and fragment span embedding at decoding time frame $t$, respectively. For fragment span $f_{i:j},( j \ge i)$ at decoding time step $t$, we obtain its span embedding from three sources: boundary tokens, inner tokens and fragment length. Specifically, we deploy a LSTM network for computing the hidden state of inner tokens, and use a looking-up table to encode the fragment length and entity type. The representation of fragment $f_{i:j}$ can be computed as follows:
\begin{equation}\label{eq5}
\begin{aligned}
&\boldsymbol{e}_{i:j}^{inner}=LSTM(\boldsymbol{h}_{t-1}^D(i),...,\boldsymbol{h}_{t-1}^D(j)) \\
&\boldsymbol{e}_{i:j}^{len}=Emb(j-i)\\
&\boldsymbol{e}_{i:j}^{boundary}=\boldsymbol{h}_{t-1}^D(i)+\boldsymbol{h}_{t-1}^D(j)\\
&\boldsymbol{w}^{f_{i:j}}_t= \boldsymbol{e}_{i:j}^{inner}+\boldsymbol{e}_{i:j}^{len}+\boldsymbol{e}_{i:j}^{boundary}\\
&\boldsymbol{w}_t=[\boldsymbol{w}^{f_{i:j}}_t;\boldsymbol{w}^{type}_t]\\
&t\in\{2,3\}
\end{aligned}
\end{equation}

\subsubsection{Scratchpad Attention Mechanism }
\begin{figure}[t]
	\centering
	\includegraphics[width=0.95\columnwidth]{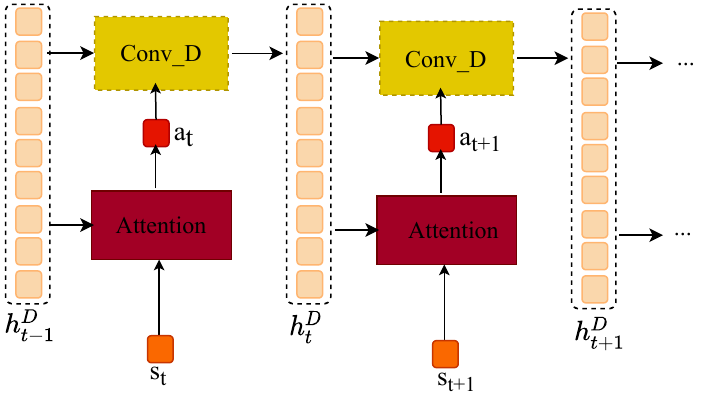}
	\caption{The workflow of the scratchpad attention mechanism: $\boldsymbol{h}_{t-1}^D$ is used to calculate attention score, and update the encoder states according to Eq.\ref{eq7} at every decoding step.} \label{fig3}
\end{figure}

Scratchpad attention mechanism was first introduced into Seq2Seq neural network architecture by~\cite{benmalek2019keeping}, to keep track of what has been generated so far and thereby guide future generation. 
Specifically, scratchpad mechanism adds one simple step to the decoder: regarding the encoder output states, $[\boldsymbol{c}_i,...,\boldsymbol{c}_n]$ (i.e., $\boldsymbol{h}_0^D$), as a scratchpad, thus it writes to them as if the set of states were external memory. Exactly how this is done is described next and shown in Figure~\ref{fig3}.

At every decoding step $t$, an attention is derived from the last updated encoder states $\boldsymbol{h}_{t-1}^D$ and the current decoder hidden state $\boldsymbol{s}_t$, to obtain attentional context-aware embedding:
\begin{equation}\label{eq6}
\boldsymbol{a}_t=\mathrm{Attention}(\boldsymbol{h}_{t-1}^D,\boldsymbol{s}_{t})
\end{equation}

Then the context-aware embedding $\boldsymbol{a}_t$ is concatenated with the last updated encoder states $\boldsymbol{h}_{t-1}^D$, and this concatenation is fed into a convolution network layer to provide new encoder states $\boldsymbol{h}_{t}^D$ to the decoder at each decoding time step $t$:
\begin{equation}\label{eq7}
\boldsymbol{h}_t^D=\mathrm{Conv\_D}([\boldsymbol{a}_{t};\boldsymbol{h}_{t-1}^D])
\end{equation}

\subsubsection{Biaffine-Attention Based Fragment Detector}
As is well known, all fragments of discontinuous entities have the same entity type, and for nested NER, we just only need to recognize the type and span of the first fragment, because all entities in nested NER datasets only have one fragment (entities are continuous). Thus, it is more important recognize the type and span of the first fragment for various complex NER.
In this paper, we adopt a biaffine attention network to extract fragment span and fragment type jointly, at each decoding time step $t$.

Considering the context of the head and tail word of fragments is different. Firstly, we encode each $\boldsymbol{h}_{t,i}^D$ via two separate MLPs to generate head/tail-sensitive word representation $\boldsymbol{h}^{s}_{t,i}$ and $\boldsymbol{h}^{e}_{t,i}$. Then, we apply the biaffine attention, as follows:
\begin{equation}\label{eq8}
\begin{aligned}
&\boldsymbol{h}^{s}_{t,i}=\mathrm{MLPs}^s(\boldsymbol{h}_{t,i}^D) \\
&\boldsymbol{h}^{e}_{t,i}=\mathrm{MLPs}^e(\boldsymbol{h}_{t,i}^D) \\
&biaff(i,k,j)=(\boldsymbol{h}^{e}_{t,j})^T\boldsymbol{U}_k\boldsymbol{h}^{s}_{t,i}+\boldsymbol{V}_k\boldsymbol{h}^{s}_{t,i}
\end{aligned}
\end{equation}
where $t$ represents the decoding step, $i$ represents the $i$-th word representation, $biaff(i,k,j)$ indicates the score of word pair $(w_i,w_j)$ holding $k$-th entity type,  the weight matrix $\boldsymbol{U}_k$ and $\boldsymbol{V}_k$ are trainable biaffine attention parameters for $k$-th entity type.
 
 Finally, we employ $sigmoid$ function on the $biaff(i,k,j)$, yielding $p(i,k,j)$ which represents the probability of the existence of $k$-th entity type between $i$-th and $j$-th word. 
 \begin{equation}\label{eq9}
 p(i,k,j)=sigmoid(biaff(i,k,j))
 \end{equation}
 
 We use the binary cross entropy loss here as the fragment span detection loss at the decoding time step $t$, denoted as $\mathcal{L}^{span_{t}}$:
 \begin{equation}\label{eq11}
 \begin{aligned}
 &\mathcal{L}^{span_{t}}=-\sum_{k=1}^K\sum_{i=1}^n\sum_{j=1}^n[y(i,k,j)logp(i,k,j)+\\
 &(1-y(i,k,j))log(1-p(i,k,j))]
 \end{aligned}
 \end{equation} 
 Where $K$ is the number of entity types, $y(i,k,j) \in \{0,1\}$, and $y(i,k,j)=1$ denotes the fact that the word pair ($w_i$,$w_j$) have the  $k$-th entity type. 

\subsection{Training Objective}
Note that, nested NER ans discontinuous NER have different loss function formulation, due to their different decoding length (as described in the \nameref{decoder} ). For nested NER datasets ($t==1$): $\mathcal{L}=\mathcal{L}^{span_1}$, and for discontinuous NER datasets ($t \in \{1,2,3\}$): $\mathcal{L}=\mathcal{L}^{span_1}+\mathcal{L}^{span_2}+\mathcal{L}^{span_3}$.
\section{Experiments}
\begin{table*}[t]
	\centering
	\setlength\tabcolsep{2pt}
	\small
	\begin{tabular}{c| c c c|c c c|c c c |c c c|c c c}
		\hline
		\multirow{2}{*}{Dataset}& \multicolumn{3}{c|}{CADEC}  &\multicolumn{3}{c|}{ShARe 13}  &\multicolumn{3}{c|}{ShARe 14}  & \multicolumn{3}{c|}{GENIA} & \multicolumn{3}{c}{ACE05} \\ 
		\cline{2-16} 
		&train & dev& test&train &dev& test &train& dev & test &train &dev& test&train&dev&test \\
		\hline
		\#Sentence& 5,340&  1,097 & 1,160 &8,505& 1,250& 9,009 & 17,407&1,361&15,850 &15,022 &1,699 & 1,855 & 7,285&968&1,058\\
		\#Entities& 4,430&   898 & 990 &5,146& 669& 5,333 & 10,354&771&7,922&47,006 &4,461 & 5,596& 24,700&3,128&3,029\\
		\%OE& 15& 14 & 13 & 7 &7& 6& 6& 7&5&18 &18 &22 & 40 & 37&39\\
		\%DE& 11& 10 & 9 & 11&11& 8& 10 & 10&7&0 &0 &0& 0 & 0&0\\
		\hline
		\#2 components & \multicolumn{3}{c|}{650}  &\multicolumn{3}{c|}{1,026}  &\multicolumn{3}{c|}{1,574}  & \multicolumn{3}{c|}{0} & \multicolumn{3}{c}{0} \\ 
			\#3 components & \multicolumn{3}{c|}{27}  &\multicolumn{3}{c|}{62}  &\multicolumn{3}{c|}{76}  & \multicolumn{3}{c|}{0} & \multicolumn{3}{c}{0} \\ 
				\#4 components & \multicolumn{3}{c|}{2}  &\multicolumn{3}{c|}{0}  &\multicolumn{3}{c|}{0}  & \multicolumn{3}{c|}{0} & \multicolumn{3}{c}{0} \\ 
		\hline
	\end{tabular}
	\caption{Statistics of datasets. `\#' denotes the amount, `\%' denotes the percentage. 'OE' denotes the overlapping entities in total entities, 'DE' denotes the discontinuous entities in the total entities.}
	\label{tab1}
\end{table*}

\subsection{Datasets and Metrics}

Following previous work~\cite{dai2020effective,hang2021a,li2021span}, to evaluate our model for recognizing nested, overlapping and discontinuous entities, we conduct experiments on three discontinuous NER datasets (\textbf{CADEC}\footnote[4]{https://data.csiro.au/collections/collection/CI10948v00}~\cite{karimi2015cadec}, \textbf{ShARe 13}\footnote[5]{https://physionet.org/content/shareclefehealth2013/1.0/}~\cite{pradhan2013task} and \textbf{ShARe 14}\footnote[6]{https://physionet.org/content/shareclefehealth2014task2/1.0/}~\cite{mowery2014task}) and two nested NER datasets (\textbf{GENIA}\footnote[7]{http://www.nactem.ac.uk/tsujii/GENIA/ERtask/report.html}~\cite{kim2003genia} and \textbf{ACE05}\footnote[8]{https://catalog.ldc.upenn.edu/ LDC2006T06}~\cite{doddington2004automatic}). Following~\cite{dai2020effective}, we preprocess the original data of CADEC by only considering the entities with Adverse Drug Event (ADE) type, because only the ADEs involve discontinuous annotations. For GENIA and ACE05 datasets, we use the same train/dev/test splits following the prior work~\cite{lu2015joint}, and both datasets contain a proportion of nested entities and have multiple entity types. The descriptive statistics of all datasets are listed in Table \ref{tab1}.

We employ the precision (P), recall (R) and F1-score (F1) as evaluation metrics. An entity is consider correct only if both the entity type and the entity boundary are correct. For a discontinuous entity, each fragment span should mach a span of the gold entity.

\subsection{Experiments Settings and Baselines}
For fair, we compare our model with several Seq2Seq-based and span-based models on both discontinuous NER and nested NER on the same runtime environment.
Limited by the space, please refer to Appendix \ref{sec:appendixA} for hyper-parameters and other settings, and the detail of the baseline methods which are compared with ours, can be refer to \ref{sec:appendixB}.

\subsection{Results on Discontinuous NER Datasets}
\begin{table*}[t]
	\centering
	\small
	\setlength\tabcolsep{3pt}
		\begin{tabular}{c|c|c c c|c c c|c c c}
			\hline
			\multirow{2}{*}{Related Work} &\multirow{2}{*}{Method}& \multicolumn{3}{c|} {CADEC}  &  \multicolumn{3}{c|}{ShARe 13} &\multicolumn{3}{c}{ShARe 14}\\ 
			&& Prec. & Rec. & F1 & Prec. & Rec. & F1 & Prec. & Rec. & F1\\ 
			\hline
			~\cite{metke2016concept}&labeling-based,CRF,BIOHD& 68.7 &66.1 &67.4 &77.0&72.9 &74.9 & 74.9 &78.5 &76.6\\
		    ~\cite{dai2020effective}$^{*}$&Transition-based, BERT& 68.8 &67.3 &68.0 &77.3&72.9 &75.0 & 76.0 &78.6 &77.3\\
		    ~\cite{hang2021a}&Seq2Seq-based, BART& 70.1 &71.2 &70.6&82.1&77.4 &79.7 & 77.2&83.7 &80.3\\
		    ~\cite{li2021span}$^{*}$ &Span-based, BERT&69.8 &68.7 &69.2 &80.6&74.2&77.3 & 76.5 &82.3&79.3\\
			\hline
			\textbf{S2F-NER (ours)} &Seq2Forest, BERT&70.5 &70.8 &\textbf{70.6}&80.5&81.7&\textbf{81.1}&80.7& 81.2 &\textbf{80.9}\\
			\hline
	\end{tabular}
	\caption{Results on the discontinuous NER datasts.  $*$ indicates we rerun their code. Note that, for related word ~\cite{hang2021a}, we only report the results of ``Word" entity representation since this representation achieves better performance almost in all datasets than orther two representations (i.e., ``Span" and ``BPE").}
	\label{tab2}
\end{table*}
\begin{table}[t]
	\centering
	\small
	\begin{tabular}{c c c c}
		\hline
		\multirow{2}{*}{Related Work} & {CADEC}  & {ShARe 13} &{ShARe 14}\\ 
		& F1  & F1  & F1\\ 
		\hline
		~\cite{dai2020effective}$^{*}$ &66.7/36.6 &61.1/49.0 &55.6/43.8 \\
		~\cite{hang2021a}$^{*}$& 67.1/43.1 &64.8/55.0 &65.7/52.7\\
		~\cite{li2021span}$^{*}$ &62.7/40.6&56.3/53.8 &59.9/51.6 \\
		\hline
		\textbf{S2F-NER (ours)} &\textbf{68.3}/\textbf{44.5} &\textbf{66.0}/\textbf{60.1} &\textbf{67.1}/\textbf{61.1}\\
		\hline
	\end{tabular}
	\caption{Results on discontinuous entity mentions. we use a slash (`/') separate two score, where the former is the score on sentences with at least one discontinuous entity mention, and the later is the score only considering discontinuous entity mentions.}
	\label{tab3}
\end{table}
Results in Table \ref{tab2} show the comparison between our model and other baseline methods in three discontinuous NER datasets. We can observe that: (1) Our model, S2F-NER, outperforms all other baseline methods on three datasets in term of F1 scores; (2) Labeling-based model achieves worst results on three datasets, which may be because such ``BIO"-Tagging extension-based method has the tag ambiguity issue; (3) Compared with ~\cite{dai2020effective} and ~\cite{hang2021a}, our model significantly outperforms transition-based model and is slightly better than Seq2Seq-based model. It is largely due to that both ~\cite{dai2020effective} and ~\cite{hang2021a} models suffer severely from exposure bias. Addition, ~\cite{hang2021a} benefits from the strong power of the pre-trained BART model, which has about 4 times as many parameters as our model (408M v.s. 111M). (4) Our model outperforms ~\cite{li2021span} by 1.4\%, 3.8\% and 1.6\% on CADEC, ShARe13 and ShARe14, respectively. We consider that it is because ~\cite{li2021span} utilize span-level classification which may introduces lots of redundant computing, affecting the accurateness. 

Additionally, we can also observe that our model exhibits balanced results of Precision and Recall on three datasets, while there is a significant gap between Precision and Recall in the results  of ~\cite{hang2021a} and ~\cite{li2021span} on ShARe13 and ShARe14 datasets. We conjecture the reason is that our attention-based learning and threshold-base prediction are more flexible and focus more relevant words.

To evaluate the effectiveness of our proposed model on recognizing discontinuous mentions, following the previous works ~\cite{dai2020effective,hang2021a}, we experiment on the subsets of the original test set where each sentence has at least one discontinuous mention or only discontinuous mentions are considered. The results are shown in Table~\ref{tab3}. We can see that our model significantly outperforms the best baselines by 1.2\%/1.4\%, 1.2\%/5.1\%, and 1.4\%/8.4\% on CADEC, ShARe13 and ShARe14 datasets respectively, which demonstrates the effectiveness of our model on discontinuous entity recognition. We consider that it is largely because that our model considers the interactions and dependencies between fragment spans and entity type, and our model provides memory information bu scratchpad attention.
\subsection{Results on Nested NER Datasets}
\begin{table*}[t]
	\centering
	\small
	\begin{tabular}{c|c|c c c|c c c}
		\hline
		\multirow{2}{*}{Related Work} &\multirow{2}{*}{Method}&\multicolumn{3}{c|}{GENIA}  & \multicolumn{3}{c}{ACE05} \\ 
		&& Prec. & Rec. & F1 & Prec. & Rec. & F1  \\ 
		\hline
		~\cite{strakova2019neural}&Seq2Seq-base,BERT& - &- & 78.2 &- & - &83.4 \\
		~\cite{hang2021a}&Seq2Seq-based, BART& 78.6 &79.3 & 78.9 &83.2& 86.4&84.7\\
		~\cite{luan2019general}&Span-based, BERT& - &-& 76.8 &- & - &82.9  \\
		~\cite{tan2020boundary}$^{*}$&Span-based+boundary,BERT& 79.2 &77.4& 78.3 &83.8& 83.9 &83.9  \\
		~\cite{li2021span}$^{*}$ &Span-based, BERT&78.8&76.9& 77.8 &84.0& 82.1 &83.0 \\
		\hline
		\textbf{S2F-NER (ours)} &Seq2Forest, BERT &80.3& 78.9 &\textbf{79.6}& 85.7&85.2 &\textbf{85.4} \\
		\hline
	\end{tabular}
	\caption{Results for two nested NER datasets. $*$ indicates we rerun their code. We only report the results of ~\cite{hang2021a} with ``Word" entity representation.}
	\label{tab4}
\end{table*}

\begin{table}[t]
	\centering
	\small
		\begin{tabular}{l c c }
			\hline
			Model & ShARe 13 (F1) & GENIA (F1)  \\ 
			\hline
			S2F-NER &\textbf{80.1}& \textbf{79.2}\\
			- BERT& 77.8 ($\downarrow$2.3)&76.5 ($\downarrow$2.7)\\
			-Conv &79.9 ($\downarrow$0.2) &78.7 ($\downarrow$0.5)\\
			- Scratchpad Attention & 78.4  ($\downarrow$1.7) &  (-)\\
			- biaffine attention & 79.2 ($\downarrow$0.9)  & 77.5 ($\downarrow$1.7) \\
			\hline
	\end{tabular}
	\caption{The results (F1) of ablation studies on ShARe13 (for discontinuous NER) and GENIA (for nested NER) datasets.}
	\label{tab5}
\end{table}

Table~\ref{tab4} shows the results of the GENIA and ACE05 datasets, which include only regular and overlapping entities. Because all the entities contained in these two datasets are continuous, so the decoding length of our model will be set to 1. In fact, in this case, our model can be viewed as an extraction model rather than a generation model. From the results of Table~\ref{tab4}, we can see that our model outperforms all baselines in F1 scores. We attribute the reason into two folds. First, since different entities do not share the same start-end pair, our fragment detection with start-end format can naturally solve the difficulty of expressing nested entities. Second, our model consider word-level correlations and joint span and type extraction, which provides more fine-grained semantics for entity recognition.
\subsection{Analysis}

\subsubsection{Ablation Studies}
Table~\ref{tab5} shows the results of ablation studies on ShARe13 and GENIA datasets, showing each component of our model have various degrees of contributes. Note that, for nested NER dataset (i.e., GENIA), there exits no ``scratchpad attention" component in our model, since all entities only have one fragment, thus does not need to fragment generation with scratchpad attention. We can see that there are considerable drops in F1, when replacing the BERT encoder with BiLSTM encoder. Besides, when we replace the Convolution layer with conventional MLPs layer, the F1 score slightly drops by 0.35\% in average, but when we remove the scratchpad attention, the F1 scores drop by 1.7\%, which indicates the scratchpad mechanism as the decoder being more focusing on related words than the standard seq2seq models. Removing biaffine attention mechanism results in a remarkable drop on the F1 score of nested NER while little drop on the F1 score of discontinuous NER. The reason may be that the word-level attention is essential to recognize nested entity mentions which make up a small portion of the ShARe13.
\begin{figure}[t]
	\centering
	\includegraphics[width=\columnwidth]{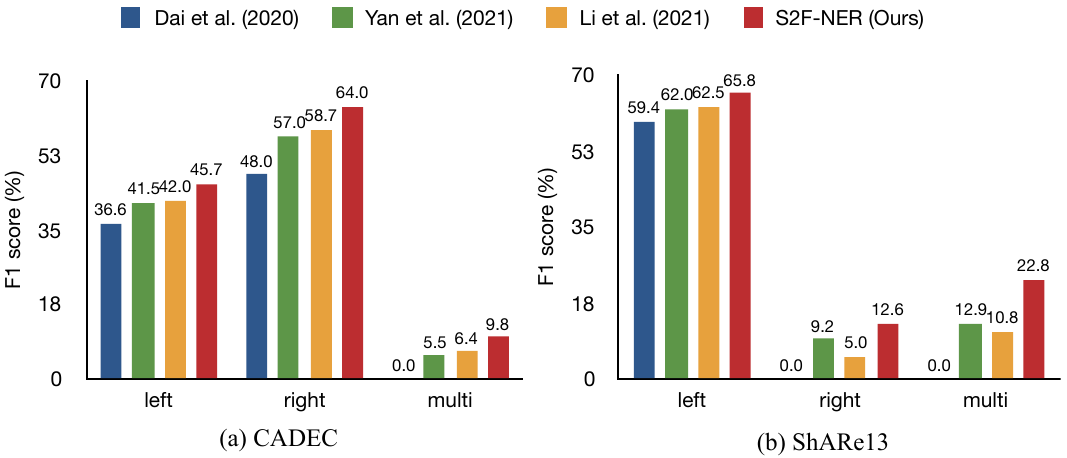}
	\caption{Effectiveness of boundary detection on CADEC and ShARe13.} \label{fig4}
\end{figure}
\subsubsection{Effectiveness for Boundary Detection }
Boundary detection is the most critical and difficult part for discontinuous NER because of the complex overlapping structures. So we explore the abilities of our model and other baselines on overlapping boundary detection by following ~\cite{dai2020effective}. Figure \ref{fig4} shows the F1 score of overlapping boundary detection on two discontinuous NER datasets with three overlapping types: left overlap, right overlap and multiple overlap\footnote[9]{Please refer to Appendix \ref{sec:appendixC} for more details.}. We can observe that our model obtains best F1 scores on different types of boundary evaluation, especially on multiple overlap, which further indicates the effectiveness of our model on discontinuous entity detection. 

\subsubsection{Inference Speed}
\begin{table}[t]
	\centering
	\small
	\begin{tabular}{l c c c }
		\hline
		Model &CADEC & ShARe 13  & GENIA   \\ 
		\hline
		~\cite{li2021span}&70.6 Sen/s & 84.1 Sen/s & 89.7 Sen/s\\
		~\cite{hang2021a} &100.4 Sen/s & 111.2 Sen/s & 121.3 Sen/s\\
		S2F-NER &\textbf{120.4} Sen/s & \textbf{150.1} Sen/s & \textbf{200.8} Sen/s\\
		\hline
	\end{tabular}
	\caption{The results (F1) of ablation studies on ShARe13 (for discontinuous NER) and GENIA (for nested NER) datasets.}
	\label{tab6}
\end{table}
To confirm that our end-to-end model equipped with sequence-to-forest paradigm is more efficient than~\cite{li2021span}(span-based) and ~\cite{hang2021a}, we make a comparison with them on inference speed. Table
\ref{tab6} reports the inference speeds of the models on three test sets. We can see that our model accelerate the decoding speed of ~\cite{li2021span} by up to 2 times, which is line with our expectation. Our model is also faster than
~\cite{hang2021a}. Since the span-pair classification of ~\cite{li2021span} introduce much redundant computation, and the full sequence of all entity spans and types of ~\cite{li2021span} leads to long decoding length. In contrast, our model avoids from the high computation cost and the long decoding length by our novel sequence-to-forest generation paradigm.

\section{Conclusion}
In this paper, we propose a unified framework for nested, overlapping, and  discontinuous NER. Unlike prior methods, we present a novel paradigm, sequence-to-forest, based on the consideration that named entity recognition is a unordered recognition task, and the observation that almost entities do not contain more than three fragments. Based on these insights, we provide a simple but effective end-to-end model, S2F-NER, which can alleviate the exposure bias issue while keeps the simplicity of Seq2Seq. Technically, our model benefits from two attention mechanisms, scratchpad attention and biaffine attention, capturing much related word semantics for NER. A series of experiments demonstrate that our model outperforms the span-based and Seq2Seq-based models on various complex NER tasks, especially for discontinuous entity recognition.

\bibliographystyle{acl_natbib}
\bibliography{mybibliography}

\clearpage
\appendix
%

\section{Experiments Settings}
\label{sec:appendixA}
We implement our model using PyTorch, and all the models are trained on a personal workstation with Intel Xeon E5 2.2GHz CPU, NVIDIA GeForce GTX 1080 Ti GPUs and 128 GB memory. The parameters of the proposed model are optimized with AdamW~\cite{loshchilov2017fixing}. The learning rate and the batch size are set to $1e{-5}$ and 20, respectively. The dimension of character embeddings and position embeddings is set to 50 and 30 respectively. The number of stacked bidirectional Transformer blocks $L$ is 12 and the size of hidden state $h^E$ is 768. The pre-trained BERT model we used are:  Yelp BERT\footnote[10]{ https://bit.ly/35RpTf0}~\cite{dai2020cost} for DEACE, and Clinical BERT\footnote[11]{https://github.com/EmilyAlsentzer/clinicalBERT}~\cite{alsentzer2019publicly} for ShARe13 and 14. We heuristically set the threshold to 0.5 for span detection. The kernel size of CNN are [3,4,5] and the filters are 200.  
We train the model at most 80 epochs and report the results on the test set using the model that has the best performance on the validation set. 

\section{Baselines}
\label{sec:appendixB}
We compare our model with several state-of-the-art models on both discontinuous NER and nested NER. 
\textbf{For discontinuous NER}, we employ the following models as baselines: (1) labeling-based~\cite{metke2016concept} expands the traditional BIO tagging scheme to seven tags and adopts classical neural sequence labeling model; (2) Transition-based~\cite{dai2020effective} generates a sequence of actions based on the shift-reduce  transition scheme; (3) Seq2Seq-Generation~\cite{hang2021a} propose a united framework for various NER subtasks with the aid of the pre-trained Seq2Seq model (BART); (4) Span-based~\cite{li2021span} utilizes a multi-task learning framework to recognizing various entities based on spans enumeration. \textbf{For nested NER}, we also make comparisons with other baselines: (1) Seq2Seq-generation based~\cite{strakova2019neural}, and (2) span-based~\cite{luan2019general,tan2020boundary} models. 

\section{Overlapping types of Boundary}
\label{sec:appendixC}
\begin{figure}[t]
	\centering
	\includegraphics[width=0.95\columnwidth]{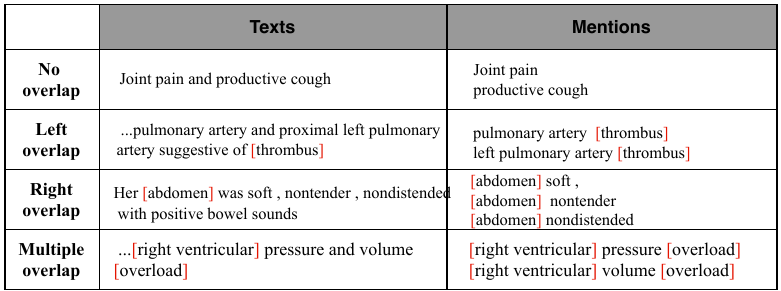}
	\caption{Examples of the overlapping patterns.} \label{fig5}
\end{figure}

following ~\cite{dai2020effective}, we classify the test set into four classes: (1) no overlap, (2) left overlap, (3) right overlap and (4) multiple overlap. Since we want to explore the ability of model on boundary detection for discontinuous entities, thus we only conducted experiments on all overlapped subsets except for no overlap set. Table~\ref{tab7} shows the statistics of overlapping patterns and Figure~\ref{fig5} gives examples for each pattern.

\begin{table}[t]
	\centering
	\small
	\begin{tabular}{c|c c c|c c c}
		\hline
		\multirow{2}{*}{Pattern} &\multicolumn{3}{c|}{CADEC}  & \multicolumn{3}{c}{ShARe13} \\ 
		&  train& dev & test & train & dev & test  \\ 
		\hline
		No &57&9& 16 &348 & 41 &193\\
		Left &270&54& 41 &167 & 11 &200\\
		Right &113&16& 23 &48 & 19 &35\\
		Multi. &51&15& 14 &18& 0 &8\\
		\hline
	\end{tabular}
	\caption{The statistics of overlapping patterns.}
	\label{tab7}
\end{table}

\end{document}